\documentclass[runningheads]{llncs}

\usepackage{eccv}

\usepackage{array}
\newcolumntype{P}[1]{>{\centering\arraybackslash}p{#1}}
\usepackage{eccvabbrv}
\usepackage{arydshln}
\usepackage{enumitem}

\usepackage{graphicx}
\usepackage{booktabs}
\usepackage{multirow}
\usepackage[accsupp]{axessibility}  
\usepackage{placeins}
\usepackage{adjustbox}

\usepackage{hyperref}

\usepackage{orcidlink}
\newcommand{\extendeddatafigs}{%
\renewcommand{\figurename}{Supplementary Fig.}
  \setcounter{figure}{0}
  \let\oldthefigure\thefigure
  
  \renewcommand{\thefigure}{\oldthefigure}
  \let\oldchapter\chapter
  \renewcommand{\chapter}{
    \let\thefigure\oldthefigure
    \let\chapter\oldchapter
    \oldchapter
}
}
\newcommand{\extendeddatatab}{%
\renewcommand{\tablename}{Supplementary Table}
  \setcounter{table}{0}
  \let\oldthetable\thetable
  \renewcommand{\thetable}{\oldthetable}
  \let\oldchapter\chapter
  \renewcommand{\chapter}{
    \let\thetable\oldthetable
    \let\chapter\oldchapter
    \oldchapter
}
}

\begin{document}


\title{Mining Field Data for \\ Tree Species Recognition at Scale} 



\author{Dimitri Gominski\inst{1}\orcidlink{0000-0002-8135-1341} \and
Daniel Ortiz-Gonzalo\inst{1}\orcidlink{0000-0002-5517-1785} \and
Martin Brandt\inst{1}\orcidlink{0000-0001-9531-1239} \and \\
Maurice Mugabowindekwe\inst{1}\orcidlink{0000-0003-0803-7417} \and
Rasmus Fensholt\inst{1}\orcidlink{0000-0003-3067-4527}
}

\authorrunning{D.~Gominski et al.}

\institute{University of Copenhagen, Denmark \\
\email{\{dg,gonzalo\}@ign.ku.dk}}

\newcommand{\nn}[1]{\ensuremath{\text{NN}_{#1}}\xspace}

\newcommand{\prm}[1]{_{#1}}
\newcommand{\dime}[1]{(#1)}

\def\l1{\ensuremath{\ell_1}\xspace}
\def\l2{\ensuremath{\ell_2}\xspace}

\newcommand*\OK{\ding{51}}

\newenvironment{narrow}[1][1pt]
	{\setlength{\tabcolsep}{#1}}
	{\setlength{\tabcolsep}{6pt}}

\newcommand{\comment} [1]{{\color{orange} \Comment     #1}} 


\newcommand{\head}[1]{{\smallskip\noindent\bf #1}}
\newcommand{\eq}[1]{(\ref{eq:#1})\xspace}

\newcommand{\red}[1]{{\color{red}{#1}}}
\newcommand{\blue}[1]{{\color{blue}{#1}}}
\newcommand{\green}[1]{{\color{green}{#1}}}
\newcommand{\gray}[1]{{\color{gray}{#1}}}


\newcommand{\tran}{^\top}
\newcommand{\mtran}{^{-\top}}
\newcommand{\zcol}{\mathbf{0}}
\newcommand{\zrow}{\zcol\tran}

\newcommand{\ind}{\mathbbm{1}}
\newcommand{\expect}{\mathbb{E}}
\newcommand{\nat}{\mathbb{N}}
\newcommand{\zahl}{\mathbb{Z}}
\newcommand{\real}{\mathbb{R}}
\newcommand{\proj}{\mathbb{P}}
\newcommand{\prob}{\mathbf{Pr}}

\newcommand{\mif}{\textrm{if }}
\newcommand{\other}{\textrm{otherwise}}
\newcommand{\minimize}{\textrm{minimize }}
\newcommand{\maximize}{\textrm{maximize }}
\newcommand{\st}{\textrm{subject to }}

\newcommand{\id}{\operatorname{id}}
\newcommand{\const}{\operatorname{const}}
\newcommand{\sgn}{\operatorname{sgn}}
\newcommand{\var}{\operatorname{Var}}
\newcommand{\mean}{\operatorname{mean}}
\newcommand{\trace}{\operatorname{tr}}
\newcommand{\diag}{\operatorname{diag}}
\newcommand{\vect}{\operatorname{vec}}
\newcommand{\cov}{\operatorname{cov}}

\newcommand{\softmax}{\operatorname{softmax}}
\newcommand{\clip}{\operatorname{clip}}

\newcommand{\defn}{\mathrel{:=}}
\newcommand{\peq}{\mathrel{+\!=}}
\newcommand{\meq}{\mathrel{-\!=}}

\newcommand{\floor}[1]{\left\lfloor{#1}\right\rfloor}
\newcommand{\ceil}[1]{\left\lceil{#1}\right\rceil}
\newcommand{\inner}[1]{\left\langle{#1}\right\rangle}
\newcommand{\norm}[1]{\left\|{#1}\right\|}
\newcommand{\frob}[1]{\norm{#1}_F}
\newcommand{\card}[1]{\left|{#1}\right|\xspace}
\newcommand{\diff}{\mathrm{d}}
\newcommand{\der}[3][]{\frac{d^{#1}#2}{d#3^{#1}}}
\newcommand{\pder}[3][]{\frac{\partial^{#1}{#2}}{\partial{#3^{#1}}}}
\newcommand{\ipder}[3][]{\partial^{#1}{#2}/\partial{#3^{#1}}}
\newcommand{\dder}[3]{\frac{\partial^2{#1}}{\partial{#2}\partial{#3}}}

\newcommand{\wb}[1]{\overline{#1}}
\newcommand{\wt}[1]{\widetilde{#1}}

\def\xssp{\hspace{1pt}}
\def\ssp{\hspace{3pt}}
\def\msp{\hspace{5pt}}
\def\lsp{\hspace{12pt}}

\newcommand{\cA}{\mathcal{A}}
\newcommand{\cB}{\mathcal{B}}
\newcommand{\cC}{\mathcal{C}}
\newcommand{\cD}{\mathcal{D}}
\newcommand{\cE}{\mathcal{E}}
\newcommand{\cF}{\mathcal{F}}
\newcommand{\cG}{\mathcal{G}}
\newcommand{\cH}{\mathcal{H}}
\newcommand{\cI}{\mathcal{I}}
\newcommand{\cJ}{\mathcal{J}}
\newcommand{\cK}{\mathcal{K}}
\newcommand{\cL}{\mathcal{L}}
\newcommand{\cM}{\mathcal{M}}
\newcommand{\cN}{\mathcal{N}}
\newcommand{\cO}{\mathcal{O}}
\newcommand{\cP}{\mathcal{P}}
\newcommand{\cQ}{\mathcal{Q}}
\newcommand{\cR}{\mathcal{R}}
\newcommand{\cS}{\mathcal{S}}
\newcommand{\cT}{\mathcal{T}}
\newcommand{\cU}{\mathcal{U}}
\newcommand{\cV}{\mathcal{V}}
\newcommand{\cW}{\mathcal{W}}
\newcommand{\cX}{\mathcal{X}}
\newcommand{\cY}{\mathcal{Y}}
\newcommand{\cZ}{\mathcal{Z}}

\newcommand{\vA}{\mathbf{A}}
\newcommand{\vB}{\mathbf{B}}
\newcommand{\vC}{\mathbf{C}}
\newcommand{\vD}{\mathbf{D}}
\newcommand{\vE}{\mathbf{E}}
\newcommand{\vF}{\mathbf{F}}
\newcommand{\vG}{\mathbf{G}}
\newcommand{\vH}{\mathbf{H}}
\newcommand{\vI}{\mathbf{I}}
\newcommand{\vJ}{\mathbf{J}}
\newcommand{\vK}{\mathbf{K}}
\newcommand{\vL}{\mathbf{L}}
\newcommand{\vM}{\mathbf{M}}
\newcommand{\vN}{\mathbf{N}}
\newcommand{\vO}{\mathbf{O}}
\newcommand{\vP}{\mathbf{P}}
\newcommand{\vQ}{\mathbf{Q}}
\newcommand{\vR}{\mathbf{R}}
\newcommand{\vS}{\mathbf{S}}
\newcommand{\vT}{\mathbf{T}}
\newcommand{\vU}{\mathbf{U}}
\newcommand{\vV}{\mathbf{V}}
\newcommand{\vW}{\mathbf{W}}
\newcommand{\vX}{\mathbf{X}}
\newcommand{\vY}{\mathbf{Y}}
\newcommand{\vZ}{\mathbf{Z}}

\newcommand{\va}{\mathbf{a}}
\newcommand{\vb}{\mathbf{b}}
\newcommand{\vc}{\mathbf{c}}
\newcommand{\vd}{\mathbf{d}}
\newcommand{\ve}{\mathbf{e}}
\newcommand{\vf}{\mathbf{f}}
\newcommand{\vg}{\mathbf{g}}
\newcommand{\vh}{\mathbf{h}}
\newcommand{\vi}{\mathbf{i}}
\newcommand{\vj}{\mathbf{j}}
\newcommand{\vk}{\mathbf{k}}
\newcommand{\vl}{\mathbf{l}}
\newcommand{\vm}{\mathbf{m}}
\newcommand{\vn}{\mathbf{n}}
\newcommand{\vo}{\mathbf{o}}
\newcommand{\vp}{\mathbf{p}}
\newcommand{\vq}{\mathbf{q}}
\newcommand{\vr}{\mathbf{r}}
\newcommand{\Vs}{\mathbf{s}}
\newcommand{\vt}{\mathbf{t}}
\newcommand{\vu}{\mathbf{u}}
\newcommand{\vv}{\mathbf{v}}
\newcommand{\vw}{\mathbf{w}}
\newcommand{\vx}{\mathbf{x}}
\newcommand{\vy}{\mathbf{y}}
\newcommand{\vz}{\mathbf{z}}

\newcommand{\vone}{\mathbf{1}}
\newcommand{\vzero}{\mathbf{0}}

\newcommand{\valpha}{{\boldsymbol{\alpha}}}
\newcommand{\vbeta}{{\boldsymbol{\beta}}}
\newcommand{\vgamma}{{\boldsymbol{\gamma}}}
\newcommand{\vdelta}{{\boldsymbol{\delta}}}
\newcommand{\vepsilon}{{\boldsymbol{\epsilon}}}
\newcommand{\vzeta}{{\boldsymbol{\zeta}}}
\newcommand{\veta}{{\boldsymbol{\eta}}}
\newcommand{\vtheta}{{\boldsymbol{\theta}}}
\newcommand{\viota}{{\boldsymbol{\iota}}}
\newcommand{\vkappa}{{\boldsymbol{\kappa}}}
\newcommand{\vlambda}{{\boldsymbol{\lambda}}}
\newcommand{\vmu}{{\boldsymbol{\mu}}}
\newcommand{\vnu}{{\boldsymbol{\nu}}}
\newcommand{\vxi}{{\boldsymbol{\xi}}}
\newcommand{\vomikron}{{\boldsymbol{\omikron}}}
\newcommand{\vpi}{{\boldsymbol{\pi}}}
\newcommand{\vrho}{{\boldsymbol{\rho}}}
\newcommand{\vsigma}{{\boldsymbol{\sigma}}}
\newcommand{\vtau}{{\boldsymbol{\tau}}}
\newcommand{\vupsilon}{{\boldsymbol{\upsilon}}}
\newcommand{\vphi}{{\boldsymbol{\phi}}}
\newcommand{\vchi}{{\boldsymbol{\chi}}}
\newcommand{\vpsi}{{\boldsymbol{\psi}}}
\newcommand{\vomega}{{\boldsymbol{\omega}}}

\newcommand{\loc}{\text{loc}}
\newcommand{\size}{\text{size}}
\newcommand{\score}{\text{score}}
\newcommand{\countf}{\text{count}}
\newcommand{\ca}{\text{CA}}

\newcommand{\rLambda}{\mathrm{\Lambda}}
\newcommand{\rSigma}{\mathrm{\Sigma}}

\makeatletter
\DeclareRobustCommand\onedot{\futurelet\@let@token\@onedot}
\def\@onedot{\ifx\@let@token.\else.\null\fi\xspace}
\def\eg{\emph{e.g}\onedot} \def\Eg{\emph{E.g}\onedot}
\def\ie{\emph{i.e}\onedot} \def\Ie{\emph{I.e}\onedot}
\def\cf{\emph{cf}\onedot} \def\Cf{\emph{C.f}\onedot}
\def\etc{\emph{etc}\onedot} \def\vs{\emph{vs}\onedot}
\def\wrt{w.r.t\onedot} \def\dof{d.o.f\onedot}
\def\etal{\emph{et al}\onedot}
\makeatother

\maketitle

\begin{abstract} 
Individual tree species labels are particularly hard to acquire due to the expert knowledge needed and the limitations of photointerpretation. Here, we present a methodology to automatically mine species labels from public forest inventory data, using available pretrained tree detection models. We identify tree instances in aerial imagery and match them with field data with close to zero human involvement. We conduct a series of experiments on the resulting dataset, and show a beneficial effect when adding noisy or even unlabeled data points, highlighting a strong potential for large-scale individual species mapping.

\end{abstract}

\vspace{1cm}

\begin{figure*}
    \centering
    \includegraphics[trim={0cm, 0cm, 0cm, 0cm},clip,width=0.9\linewidth]{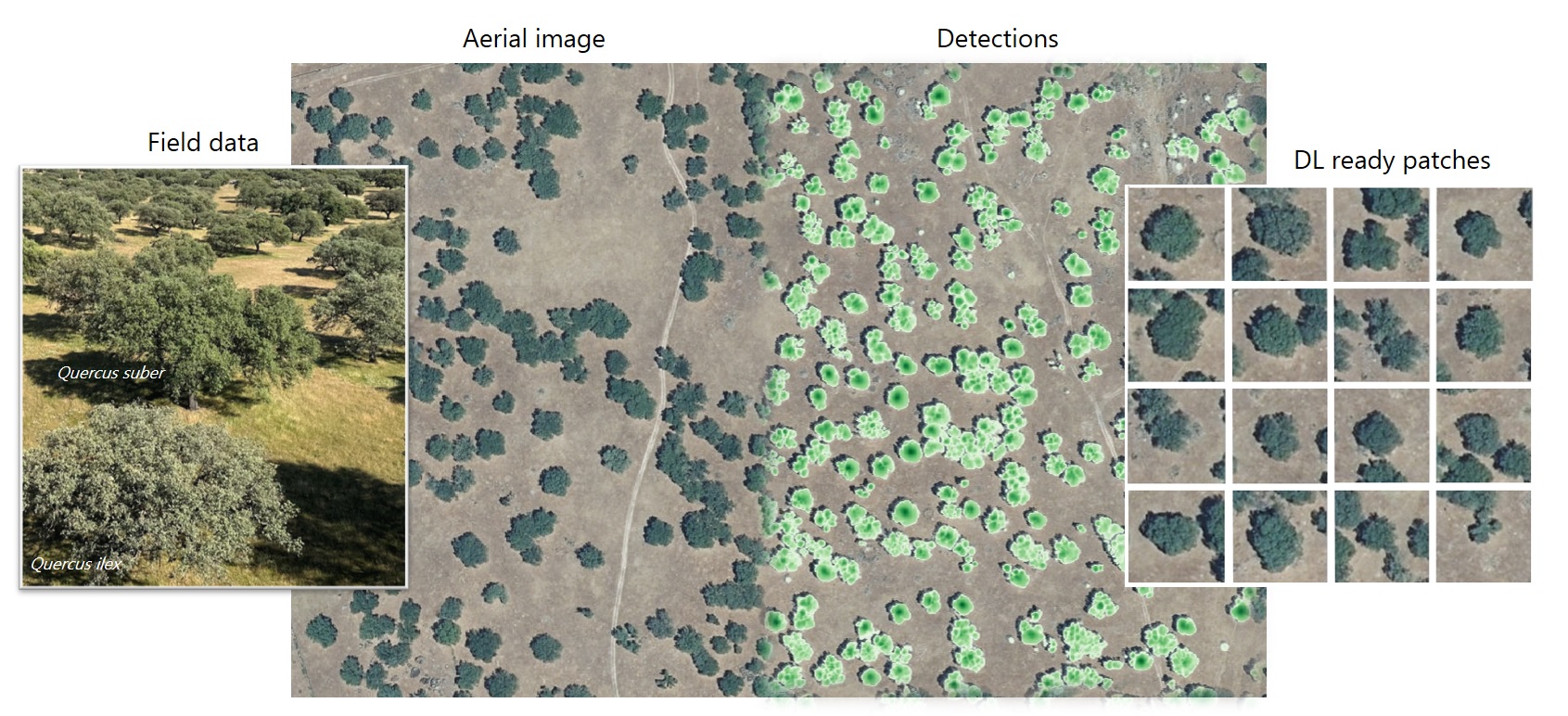}
    \caption{National forest inventories provide a standardized and extensive source of species information at individual level. We perform individual tree detection on aerial photography to extract individual positions and match them with the field data to extract deep learning-ready patches with species labels.}
    \label{fig:visualabstract}
\end{figure*}

\section{Introduction}
\label{sec:intro}

Monitoring trees is essential for addressing biodiversity loss and climate change, maintaining ecosystem services, and guiding global conservation and restoration efforts \cite{ipbes_2019, cavender-bares_integrating_2022}. Tree data serve to provide baselines, track changes over time, train and validate ecological models, as well as inform policies and decision making \cite{ferretti_advancing_2024}. While forest inventories provide high-quality tree data, such as species and ground measurements, they are labor-intensive, costly, and often limited in spatial and temporal scope \cite{ferretti_advancing_2024}. Remote sensing methods through aerial and satellite images offer broader spatiotemporal coverage for tree monitoring, but lack the precision and resolution of ground surveys \cite{ferretti_advancing_2024, cavender-bares_integrating_2022}. Recent advances in deep learning have begun to optimize this tradeoff between detail and scale by linking field and image data \cite{ gominski2023benchmarkingindividualtreemapping}. This has, for instance, enabled the leap from general tree cover assessments to individual tree counts and traits such as height or biomass using aerial and satellite images \cite{brandt_unexpectedly_2020, mugabowindekwe_nation-wide_2023, tucker_sub-continental-scale_2023}. Yet, our current capabilities fall short in identifying tree species at large scales, hindering new avenues for ecological research and environmental management.

Bridging the gap between remote sensing and terrestrial data is needed to perform species identification. While high-resolution aerial imagery is now widely accessible in many countries, species data at the individual tree level is often nonexistent or not publicly available \cite{vidal_role_2016}. National Forest Inventories (NFIs) are invaluable repositories of tree species data, meticulously collected through extensive efforts, which have evolved from focusing on productive purposes to embracing a broader range of ecological and environmental variables \cite{vidal_role_2016}. However, trees outside NFIs plots, as well as trees outside of forests such as those in agricultural and urban landscapes, are left unmonitored \cite{liu_overlooked_2023}. Furthermore, NFIs often do not provide the precise locations of the parcels (and thus individual trees) \cite{schadauer_access_2024, paivinen_ensure_2023}, which are key for matching ground-based with remotely sensed information \cite{gessler_finding_2024}.

In this study, we develop a framework for performing tree species recognition at scale using publicly available aerial images (with pixel resolution of 15-25 cm) and publicly available forest inventory data. Our approach involves three main steps: (1) mining individual trees detections in aerial images using pretrained deep models; (2) matching detected trees and NFI data to build a dataset of individual trees; and (3) training and validating a deep learning model for species recognition across forest and non-forest ecosystems.

\section{Related Work}

Recent years have seen groundbreaking advances in mapping overstory trees in satellite and aerial imagery at scale using deep learning approaches, including tree counting \cite{brandt_unexpectedly_2020}, bounding box detection \cite{weinstein_remote_2021}, tree segmentation \cite{mugabowindekwe_nation-wide_2023, freudenberg_individual_2022}, and heatmap peak detection \cite{brandt_severe_2024}. 
But identifying individual tree species can be challenging due to insufficient image resolution \cite{beloiu_individual_2023}, the variability in species traits and the structural complexity of the ecosystem \cite{fassnacht_review_2016, beloiu_individual_2023}. A number of studies have classified species using remote sensing with a certain accuracy \cite{fassnacht_review_2016, zhang_2022, beloiu_individual_2023, zhang_tree_2020, mayra_tree_2021, schiefer_mapping_2020}, with a notable effort recently on building large-scale datasets of urban tree species \cite{beery_auto_2022, BRANSON201813}, one species over large areas \cite{huang_baobab}, or monospecific forested areas \cite{gaydon2024pureforestlargescaleaeriallidar} using public data. However, to our knowledge, there is no existing dataset for individual tree species recognition "in the wild", \ie in mixed forests and trees outside of forests at scale, with no assumption of species distribution.  

A key challenge hampering individual species identification is the lack of reliable labels at individual level. While NFIs do produce individual level measurements (mostly in Europe) \cite{ferretti_advancing_2024}, there is a high uncertainty on individual positions due to the accumulation of noise from GPS measurements, image registration, and temporal mismatches. This particular situation of high volumes of data that also come with a high level of noise is not uncommon in remote sensing, and has been addressed through various works exploiting unreliable labels at large scale for canopy height mapping \cite{lang_high-resolution_2023}, individual tree detection \cite{weinstein_remote_2021}, or building mapping \cite{girard_noisy_2019}. Parallel to this line of work, recent advances in semi-supervised learning \cite{tarvainen_mean_2017, xu_end--end_2021, yang_interactive_2021} are a valuable toolbox for building label-efficient methods.

\section{Dataset Construction}
\label{sec:dataset}

We collected aerial imagery from publicly available campaigns conducted by the Spanish National Plan for Aerial Orthophotography (PNOA) \cite{PNOA}. The images have a ground sampling distance (GSD) of 15-25cm, and include four bands as RGB + NIR. We covered over 65,000 km² in the provinces of Badajoz, Cáceres, Madrid, and León, chosen for their diverse Mediterranean climates, from xeric to humid and continental, and altitudes ranging from 200 to 2,648 meters above sea level. For the field data, we made use of the 4th NFI of Spain \cite{IFN}, which collects extensive data at national-level about forests and their evolution, both from a forest management and ecological perspective. It distributes plots roughly at the intersections of a 1 km² cell grid, which are monitored every 10 years. Plots consist on circular parcels, where all trees with a trunk diameter $\geq7,5$cm are recorded in a $25$m radius. This includes identifying tree species, recording their position using local coordinates from the center of the plot, measuring functional traits such as diameter and height, and evaluating their health status among other attributes.

\subsection{Individual Tree Detection}

We use the aerial imagery and available models to perform individual tree detection. In recent years, there has been signficant progress in tree detection, from a fruitful combination of deep vision models and large-scale labeling efforts \cite{cheng_scattered_2024, li_deep_2023, brandt_unexpectedly_2020, brandt_severe_2024, freudenberg_individual_2022, weinstein_benchmark_2021}.

An ensemble of models was employed to predict individual tree positions. The ensemble includes two types of models:

\begin{itemize}
    \item Models trained for pixel-wise classification (segmentation) on rasterized polygon labels for individual crowns. We used a combination of the Tversky \cite{salehi_tversky_2017} and Focal \cite{lin_focal_2017} loss functions. The Tversky loss encourages a high overlap between predictions and labels, while the Focal loss focuses on misclassified pixels, making them complimentary.

\begin{equation}
    L_{\text{seg}} = 0.6 * L_{\text{tversky}} + 0.4 * L_{\text{focal}}
\end{equation}

    \item Models trained for individual tree detection with Gaussian heatmap modelling \cite{ventura_individual_2022}, on point labels. Each point $p$ at position $(x, y)$ generates a 2D Gaussian kernel 
    
\begin{equation}
    \mathbf{h}^p_{i, j} = \exp(-{\frac{(i - x)^2 + (j - y)^2}{2\sigma^2}})
\end{equation}

    with a fixed-size standard deviation $\sigma$. Individual Gaussian kernels are max-pooled pixel-wise to give a target heatmap, for which we optimize with a simple MSE loss

\begin{equation}
        L_{\text{hm}} = \lVert \hat{\mathbf{h}} - \mathbf{h} \lVert_2, \quad \mathbf{h} = \underset{p}{\text{max}}(\mathbf{h}^p)
\end{equation}
\end{itemize}

Segmentation models predict the extent and shape of the canopy cover, but tend to output a continuous mask in dense forests. On the other hand, heatmap-based detection models give an equal weight to each individual tree which encourages an accurate positioning and count of predictions, but can fail to capture the extent of the crown area since they train with a fixed $\sigma$. With pixel-wise ensembling, we exploit the synergy between those two approaches and get predictions that are more stable, more accurate, and that better predict crown area and center location. In both cases, we use a simple UNet \cite{ronneberger_u-net_2015} architecture with a ResNet50 \cite{he_deep_2016} backbone.

We trained models on two datasets. The first dataset contains around 100k individual polygons with aerial imagery in Rwanda \cite{mugabowindekwe_nation-wide_2023}, RGB 25cm GSD. The second dataset contains around 40k individual points with aerial imagery in France, RGB 20cm GSD. The Rwanda dataset was used with segmentation and heatmap detection  (after converting the polygons to their centroids), and the France dataset was used with heatmap detection only. Every dataset/model combination was systematically trained on three random 80/20 train/val splits. This gives a total of 9 models.

After predicting with each model separately, the 9 outputs are merged into a final raster with pixel-level arithmetic averaging. Detections are extracted as points with local maxima identification with a kernel size K=2m and a confidence threshold T=0.25. These values were manually chosen to balance over-prediction on closed canopies and under-prediction on sparse areas or smaller trees, they can also be optimized with a sweep on a validation set, if available\cite{brandt_severe_2024}.

\subsection{Matching with Field Data}

We consider a list of $N$ detections and $M$ field-measured trees on a local patch. First, we conduct 1-to-1 matching by minimizing a patch-level cost matrix, where pairwise assignment cost is the intercenter Euclidean distance:

\begin{equation}
    \label{eq:cost}
    c_{ij}= \begin{cases} 
    \lVert p_i - \hat{p_j} \rVert_2 & \text{if } \lVert p_i - \hat{p_j} \rVert_2 < 4\text{m} \\
    \infty & \text{otherwise,}
    \end{cases}
\end{equation}

with a threshold of 4m to ensure that matches stay realistic. Compared to an exhaustive assignment to the closest candidate, 1-to-1 matching better handles dense areas where the closest candidate might not be the best choice if considering a wider context. 
We will refer to matched detections as "\textbf{verified}", and unmatched detections as "\textbf{unverified}".

After matching, we differentiate two situations:

\begin{enumerate}[label=\Alph*]
    \item the parcel is monospecific, \ie only has trees of the same species. In that case, we assign the same species to all detections within the radius. Verified and unverified detections are kept.
    \item the parcel has multiple species. In that case, we only assign species to verified detections and keep them, and ignore others.
\end{enumerate}

While verified detections can be considered as "ground truth", suitable for training and validating classification models, unverified detections in monospecific plots are not as reliable. We treat these cases as noisy labels, and use them only for training.

\subsection{Statistics}

 \begin{figure}[h]
    \centering
    \includegraphics[width=0.65\linewidth]{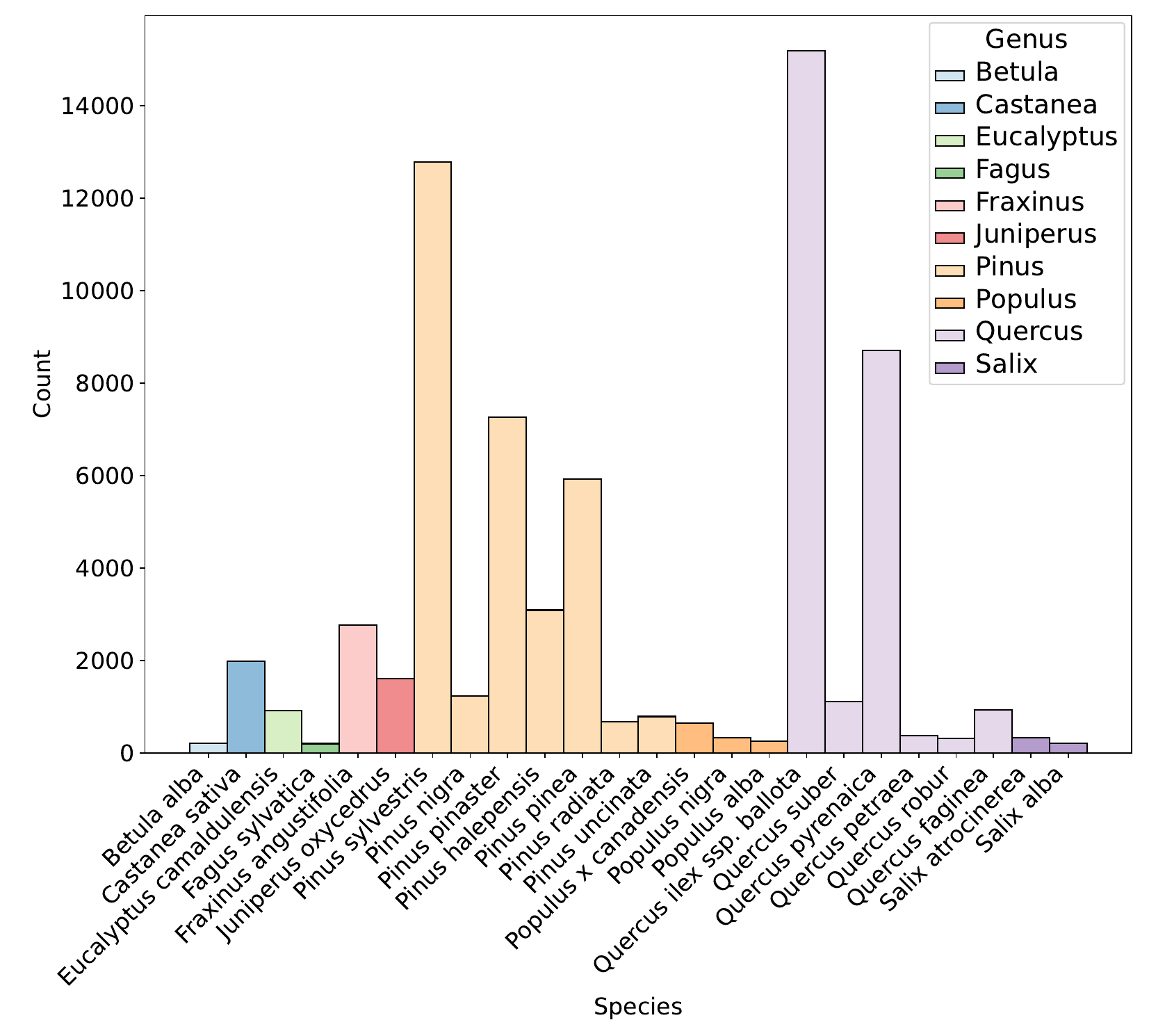}
    \caption{Our dataset has a variety of species with a typical imbalanced distribution. We only plot species with more than 200 individuals.}
    \label{fig:species_distribution}
\end{figure}

The final dataset contains 69180 individual trees, of which $20.4\%$ were matched with a field observation with a maximum distance of 4m, the remaining being detected in a monospecific parcel and assigned the corresponding label. Of the 1518 parcels, $61.6\%$ (935) are monospecific, the rest having multiple species.

The dataset contains 56 unique species belonging to 28 different genera. We plot the species histogram in Figure~\ref{fig:species_distribution}.

\subsection{Quality Control}

We asked an expert in forestry to label a random selection of 48 parcels. We compared the hand-labels with the field-measured tree positions at these parcels, and with the ensemble model predictions,  to assess A) if the detection model maintains its accuracy on the new areas we consider here, B) the level of agreement between remote observations and field measures. We conduct one-to-one matching to collect true positive, false positives, false negatives, and report the F1 score, the harmonic mean of precision and recall.

We indicate results in Table~\ref{tab:qualitycontrol}. We note a high agreement between manual labels and model predictions, and a low agreement of the latter two with field measures. Notably, field data reports a significantly lower count of trees overall. This mismatch highlights the level of uncertainty on tree positions, due to inconsistencies between NFI standards (only measuring trees above a certain radius and within a certain area, instrument and human errors) and photointerpretation from remote sensing data (only using the visible crown to identify trees, data and model errors).

Most interestingly perhaps, we note that the model predictions agree more with the field data than human labels do. In other words, the performance of our ensemble deep model surpasses that of a human annotator, if we consider the field data as close to being "ground truth". This encourages us to consider model predictions as a good source of tree positions for mining species labels.

\begin{table*}
\caption{Agreement between detections from aerial photography vs. manual labels from photointerpretation vs field-reported tree positions.}
\label{tab:qualitycontrol}
\centering
  \begin{tabular}{ll|P{2cm}P{2cm}P{2cm}}
    \toprule
     Reference & Matched & Difference & F1 & Avg. distance\\ \hline
     Field data & Manual labels & $+1342$ & 41.4 & 1.2m\\
     Field data & Predictions & $+1099$ & 47.9 & 1.1m\\
     Manual labels & Predictions & $-243$ & 76.2 & 0.6m\\
    \bottomrule
\end{tabular}
\end{table*}

\section{Experiments}
\label{sec:experiments}

We trained a simple ResNet34 classifier on our dataset. We conduct experiments to assess the potential of automatically mining tree species from monospecific plots. We report three metrics: overall accuracy (OA), class-averaged intersection over union (mIoU), and class-averaged recall (AR). While OA indicates how well the model predicts species overall, class-averaged metrics remove the bias towards frequent species.

We conduct three experiments (Table~\ref{tab:results}): training only on verified labels, training on verified and unverified labels, and training on all labels + a volume of unlabeled patches. For this last setup, we apply a recent semi-supervised approach, deep label propagation \cite{iscen_label_2019}, which builds a similarity graph between labeled and unlabeled data points to extract pseudo-labels. We experiment with 50k (around $50\%$ labeled data) and 500k (around $10\%$ labeled data) additional unlabeled patches that we randomly extracted from areas that did not contain NFI parcels.

\begin{table*}[t!]
\caption{Species classification experiments. Progressively adding noisy and unlabeled data enhances performance.}
\label{tab:results}
\centering
  \begin{tabular}{l|P{2cm}P{2cm}P{2cm}}
    \toprule
    & OA & mIoU & AR\\
    \hline
    Only verified & $40.3\scriptstyle{\pm2.5}$ & $12.9\scriptstyle{\pm1.6}$ & $24.8\scriptstyle{\pm4.1}$\\
    + unverified & $44.4\scriptstyle{\pm1.5}$ & $16.8\scriptstyle{\pm1.5}$ & $31.2\scriptstyle{\pm2.9}$\\
    + 50k unlabeled & $45.5\scriptstyle{\pm0.7}$ & $23.3\scriptstyle{\pm0.2}$ & $42.6\scriptstyle{\pm0.4}$\\
    + 500k unlabeled & $54.9\scriptstyle{\pm1.2}$ & $29.7\scriptstyle{\pm1.2}$ & $54.8\scriptstyle{\pm2.4}$\\
    \bottomrule
\end{tabular}
\end{table*}

We note a significant and clear positive effect when adding first unverified patches, then unlabeled patches. Class-averaged metrics in particular were increased by a factor of 2 between a model trained only on verified patches and a model trained with only 10\% of labeled data, indicating that unlabeled patches contribute to building more balanced models, that do not ignore less-represented species.

\section{Conclusion}

We introduced a pipeline to automatically match field observations with predictions on aerial photography from an ensemble of pretrained models for individual tree detection and segmentation. Comparing model predictions with hand-labels, we noted that the ensemble surpasses human performance, a strong signal for using predictions at large scale. Importantly, the pretrained models were \textbf{not} fine-tuned on our target area, making ensembling a promising strategy for generalizable tree detection.

Our experiments on the automatically generated dataset of tree species show that a classifier is able to learn from this not-so-reliable data, and even benefits from large volumes of unverified or unlabeled examples. While forest inventory data can be unreliable and difficult to work with from a remote sensing perspective, we show here that it is possible to harness the sheer volume of data with modern deep learning methods and learn balanced species classifiers. Note that this strategy can be effortlessly generalized to estimate biomass, height, health, and other attributes commonly collected in forest inventories.

%
%
\bibliographystyle{splncs04}
\bibliography{main}

\end{document}